\pdfoutput=1

\documentclass[11pt]{article}

\usepackage{acl}

\usepackage{times}
\usepackage{latexsym}

\usepackage[T1]{fontenc}

\usepackage[utf8]{inputenc}

\usepackage{microtype}

\usepackage{inconsolata}

\usepackage{graphicx}
\usepackage{multicol}
\usepackage{multirow}
\usepackage{makecell}
\usepackage{booktabs} 
\usepackage{amssymb}
\usepackage{amsmath}
\usepackage{subfigure}
\usepackage{longtable}
\usepackage{enumitem}
\usepackage{tabularx}
\usepackage{arydshln}
\usepackage{hyperref}
\title{MedINST: Meta Dataset of Biomedical Instructions}

\author{
    \textbf{Wenhan Han\textsuperscript{1}},
    \textbf{Meng Fang\textsuperscript{2,1}},
    \textbf{Zihan Zhang\textsuperscript{3}},
    \textbf{Yu Yin\textsuperscript{2}},
    \\
    \textbf{Zirui Song\textsuperscript{3}},
    \textbf{Ling Chen\textsuperscript{3}},
    \textbf{Mykola Pechenizkiy\textsuperscript{1}},
    \textbf{Qingyu Chen\textsuperscript{4}}
    \\
    \textsuperscript{1}Eindhoven University of Technology
    \textsuperscript{2}University of Liverpool
    \\
    \textsuperscript{3}University of Technology Sydney
    \textsuperscript{4}Yale University
    \\
\texttt{w.han@tue.nl, Meng.Fang@liverpool.ac.uk, qingyu.chen@yale.edu}
}

\begin{document}
\maketitle
\begin{abstract}
The integration of large language model (LLM) techniques in the field of medical analysis has brought about significant advancements, yet the scarcity of large, diverse, and well-annotated datasets remains a major challenge. Medical data and tasks, which vary in format, size, and other parameters, require extensive preprocessing and standardization for effective use in training LLMs. To address these challenges, we introduce \textsc{MedINST}, the Meta Dataset of Biomedical Instructions, a novel multi-domain, multi-task instructional meta-dataset. \textsc{MedINST} comprises 133 biomedical NLP tasks and over 7 million training samples, making it the most comprehensive biomedical instruction dataset to date.
Using \textsc{MedINST} as the meta dataset, we curate \textsc{MedINST32}, a challenging benchmark with different task difficulties aiming to evaluate LLMs' generalization ability.
We fine-tune several LLMs on \textsc{MedINST} and evaluate on \textsc{MedINST32}, showcasing enhanced cross-task generalization.
\end{abstract}

\section{Introduction}
Recent advancements in large language models
(LLMs), such as GPT-4 \cite{openaiGPT4TechnicalReport2024}, LLaMA-3 \cite{llama3_2024} and Mistral \cite{jiangMistral7B2023} have demonstrated impressive performance across various open-domain NLP tasks. Rather than developing specialized, task-specific systems, there is an increasing focus on rapidly adapting LLMs to specific tasks through simple prompting techniques. Studies have demonstrated that such prompted LLMs can achieve and even outperforms the capabilities of specialized models in a variety of NLP tasks \cite{radfordLanguageModelsAre, brownLanguageModelsAre2020, weiFinetunedLanguageModels2022, sanhMultitaskPromptedTraining2022}.  Due to the high cost of pre-training LLMs, instruction finetuning has become the standard method for adapting base LLMs to specific domains. Therefore, training domain-specific LLMs has largely shifted to a data-centric approach. 

In recent years, the field of medical analysis has experienced a transformative shift with the integration of large language model (LLM) techniques, fundamentally expanding the landscape of diagnostic and therapeutic strategies. The advancement of this field relies heavily on the availability of large, diverse, and well-annotated datasets, which are crucial for training robust and effective machine learning (ML) models.
Although specialized biomedical models such as BioBERT \cite{leeBioBERTPretrainedBiomedical2020}, ClinicalXLNET \cite{huangClinicalXLNetModeling2019}, BioM-Transformers \cite{alrowiliBioMTransformersBuildingLarge2021} and SciFive \cite{phanSciFiveTexttotextTransformer2021} have achieved a success, they rely on task-specific modules and follow a pre-train then fine-tune paradigm for specified tasks \cite{liuPretrainPromptPredict2021, wangPretrainedLanguageModels2023}. In this context, generalizing to unseen tasks is computationally expensive and time-consuming. Attempts exist such as In-BoXBART \cite{parmarInBoXBARTGetInstructions2022} and  BioMistral \cite{labrakBioMistralCollectionOpenSource2024} are finetuned with biomedical instructions. However, the data involved in training and evaluation are limited. Collecting raw medical data and converting it into a format suitable for LLM applications is often complex and challenging. Medical data and tasks vary significantly in format, size, and other parameters, necessitating extensive preprocessing and standardization. This task becomes even more intricate when integrating multiple datasets from various domains into a cohesive, standardized format. This raises the necessity of a comprehensive biomedical instruction meta-dataset.

\begin{table*}[h]
\centering
\small
\resizebox{\textwidth}{!}{%
\begin{tabular}{lcccccc}
\toprule
\textbf{Resource} & \textbf{\textsc{MedINST}} & \textbf{\textsc{Sup-NatInst}} \cite{wangSuperNaturalInstructionsGeneralizationDeclarative2022} & \textbf{BoX} \cite{parmarInBoXBARTGetInstructions2022}  & \textbf{BLURB} \cite{guDomainSpecificLanguageModel2021} \\
 & (this work) & (Biomedicine) & &   \\
\midrule
Has task instructions? & \checkmark & \checkmark & \checkmark & \texttimes\\
Has multi-task datasets? & \checkmark & \texttimes & \texttimes & \texttimes  \\
Has examples? & \checkmark & \checkmark & \checkmark & \texttimes  \\
Is public? & \checkmark & \checkmark & \checkmark  & \checkmark  \\ \hline
Number of tasks & 133 & 30 & 32 & 13 \\
Number of instructions & 133 & 30 & 32 & -  \\
Number of annotated task types & 12 & - & 9 & 6  \\
Avg. task definition length (words) & 45.98 & 56.6 & - & -  \\
\bottomrule
\end{tabular}
}
\caption{Comparison of \textsc{MedINST} to several datasets in biomedical field.}
\label{tab:compare}
\end{table*}

To address the problem, we release \textsc{MedINST}\footnote{The code, models and data are available at \url{https://github.com/aialt/MedINST}.}, an instruction dataset collection includes 133 biomedical NLP tasks in 12 categories such as Named Entity Recognition (NER), Question-Answering (QA), Relation Extraction (RE), etc. A benchmark is set by curate a test set from the entire collected dataset. In the experiment, multiple scales of LLMs are finetuned on our training data to demonstrate the generalization performance enhancement. Table \ref{tab:compare} presents the comparison of \textsc{MedINST} to relevant datasets in biomedical field.
In summary, our contributions are:
\setlist{nolistsep}
\begin{itemize}[noitemsep]
    \item We release a novel dataset \textsc{MedINST}, a biomedical instruction meta-dataset that involves 7M samples spanning 133 tasks among 12 categories.
    \item Using the meta dataset, we curate \textsc{MedINST32}, a challenging benchmark for evaluating the cross-task generalization ability of LLMs in the biomedical domain.
    \item We introduce instruction fine-tuned LLMs on \textsc{MedINST} based on LLaMA-3 and conduct comprehensive evaluation and analysis across multiple baselines.
\end{itemize}

\section{Related Work}
\subsection{Instruction Finetuning}
Instruction finetuning involves training models to follow specific instructions, often resulting in improved generalization and the ability to perform a wider range of tasks \cite{weiFinetunedLanguageModels2022}. There are already numerous open-domain instruction datasets and finetuned models. \textsc{Natural Instructions} is curated from samples of different NLP datasets and the crowdsourcing instructions used to annotate them. \textsc{FLAN} 2021 \cite{weiFinetunedLanguageModels2022} and 2022 \cite{longpreFlanCollectionDesigning2023} provide extensive publicly available set of tasks and methods for instruction
tuning. \textsc{FLAN} models are trained on the collection and exhibits strong generalization performance on a variety tasks. The InstructGPT \cite{ouyangTrainingLanguageModels2022} model benefits in part from a substantial dataset of prompts gathered through various synthetic data augmentation methods. However, this dataset is not publicly accessible. \textsc{Super-Natural Instructions} \citep{wangSuperNaturalInstructionsGeneralizationDeclarative2022} is established as a benchmark of 1,616 diverse NLP tasks along with expert-written instructions. The collection covers 76 distinct task types, providing a rigorous benchmarking of generalization performance of LLMs. The corresponding trained model \textsc{T$k$-Instruct} outperforms InstructGPT despite being an order of magnitude smaller. Self-Instruct \cite{wangSelfInstructAligningLanguage2023} provides a new approach for instruction fine-tuning. It involves bootstrapping off the generations of pre-trained language models to improve the instruction-following performance of themselves. After the great success of ChatGPT \cite{OpenAI_chatgpt2022}, many efforts have been made to use data generated by ChatGPT to train their own large language models (LLMs). Alpaca \cite{stanford_alpaca} is finetuned from LLaMA \cite{touvronLLaMAOpenEfficient2023} on 52k instruction-following instances generated by Text-davinci-003. Compared to open-domain instruction datasets, instruction datasets in the biomedical field are relatively scarce. MedAlpaca \cite{hanMedAlpacaOpenSourceCollection2023} utilizes a data collection of 160k entries from a reformatted medical NLP task and a crawl of internet resources. ChatDoctor \cite{liChatDoctorMedicalChat2023} is trained using 100k patient-doctor dialogues from an online medical consultation platform. Similar to Alpaca, AlpaCare \cite{zhangAlpaCareInstructiontunedLarge2024} uses medical related instruction demonstrations generated by ChatGPT to train on LLaMA. By prompting ChatGPT to conduct self-chat, Baize \cite{xuBaizeOpenSourceChat2023} collect the dialogues to train a specialized model for healthcare. Additionally, BioMistral \cite{labrakBioMistralCollectionOpenSource2024} and PMC-LLaMA \cite{wuPMCLLaMABuildingOpensource2023} use medical-related corpora to pretrain their respective base models, followed by finetuning with an instruction dataset. All these models are only finetuned on a limited number of tasks, making them prone to failure when confronted with new tasks.
Our dataset focuses on biomedical domain, offering comprehensive instruction-following demonstrates spanning 133 tasks in 12 task categories, facilitating LLMs generalizing to unseen tasks.

\subsection{Biomedical Benchmarks}
Biomedical workshops, such as BioNLP \cite{kimOverviewBioNLP092009} and BioCreative \cite{hirschmanOverviewBioCreAtIvECritical2005}, often employ task-specific benchmark datasets. With the rise of LLMs, there are higher expectations for the comprehensive capabilities of medical models. As a result, evaluating them on a single task is no longer sufficient. BLUE (Biomedical Language Understanding Evaluation) \cite{pengTransferLearningBiomedical2019} took the first step by constructing a benchmark that includes 10 datasets covering 5 different task types. Building on this foundation, BLURB (Biomedical Language Understanding and Reasoning Benchmark) \cite{guDomainSpecificLanguageModel2021} expanded the dataset to 13, encompassing 7 different types. Instruction datasets exist for few and zero-shot evaluations. \citet{agrawalLargeLanguageModels2022} introduce 3 datasets for clinical information extraction by reannotating the CASI datset. \textsc{Super-Natural Instruction} \cite{wangSuperNaturalInstructionsGeneralizationDeclarative2022} delivers 1600+ open-domain NLP tasks, among which 30 tasks are related to medicine and healthcare. Tailored for biomedicine, BoX \cite{parmarInBoXBARTGetInstructions2022} provides 32 tasks in the scope of 9 categories. BigBIO \cite{friesBigBIOFrameworkDataCentric2022} focuses on the process of constructing meta-datasets, providing unified schema for 126 existing datasets across various tasks and offering tools for building new datasets. However, it does not contain instructions and the datasets are not in text generation format. Our dataset offers an extensive instruction benchmark including 32 tasks representing a comprehensive evaluation of LLM performance in biomedical fields. 

\section{\textsc{MedINST}: Meta Dataset of Biomedical Instructions}
\begin{figure*}[h]
    \centering
    \vspace{0pt}
    \hspace{0pt}
    \subfigure[Treemap.]{
    \includegraphics[width=0.7\textwidth]{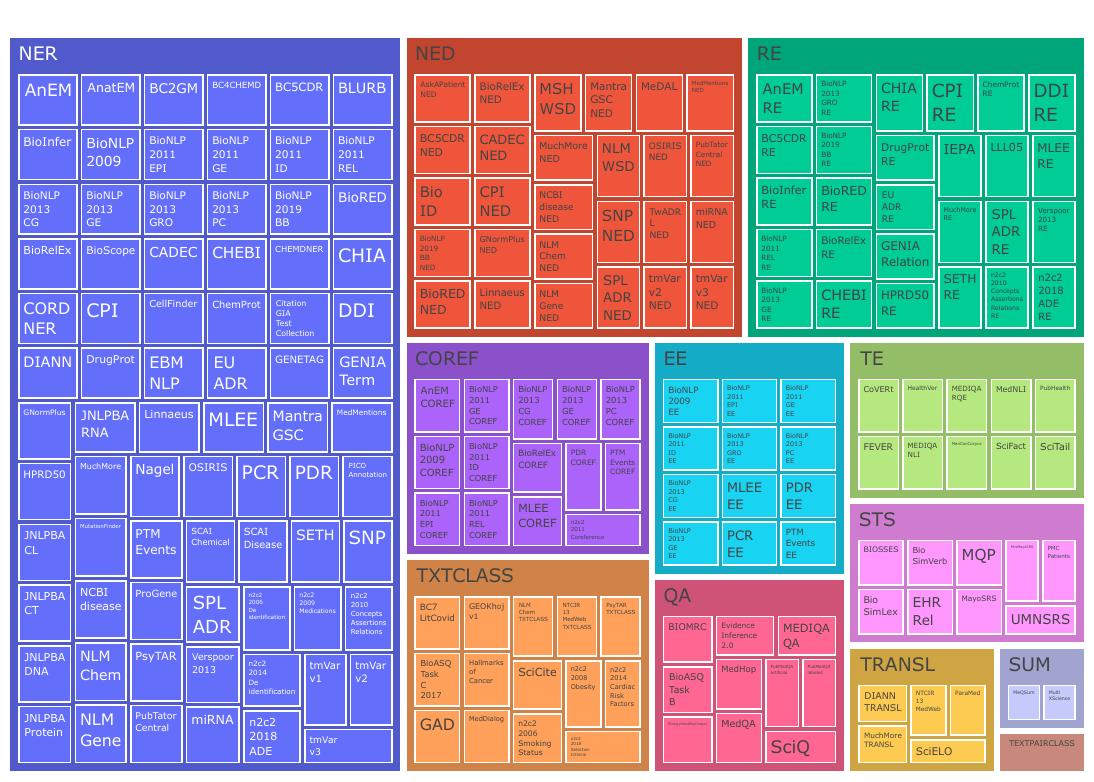}}
    \subfigure[Number of samples.]{
    \includegraphics[width=0.25\textwidth]{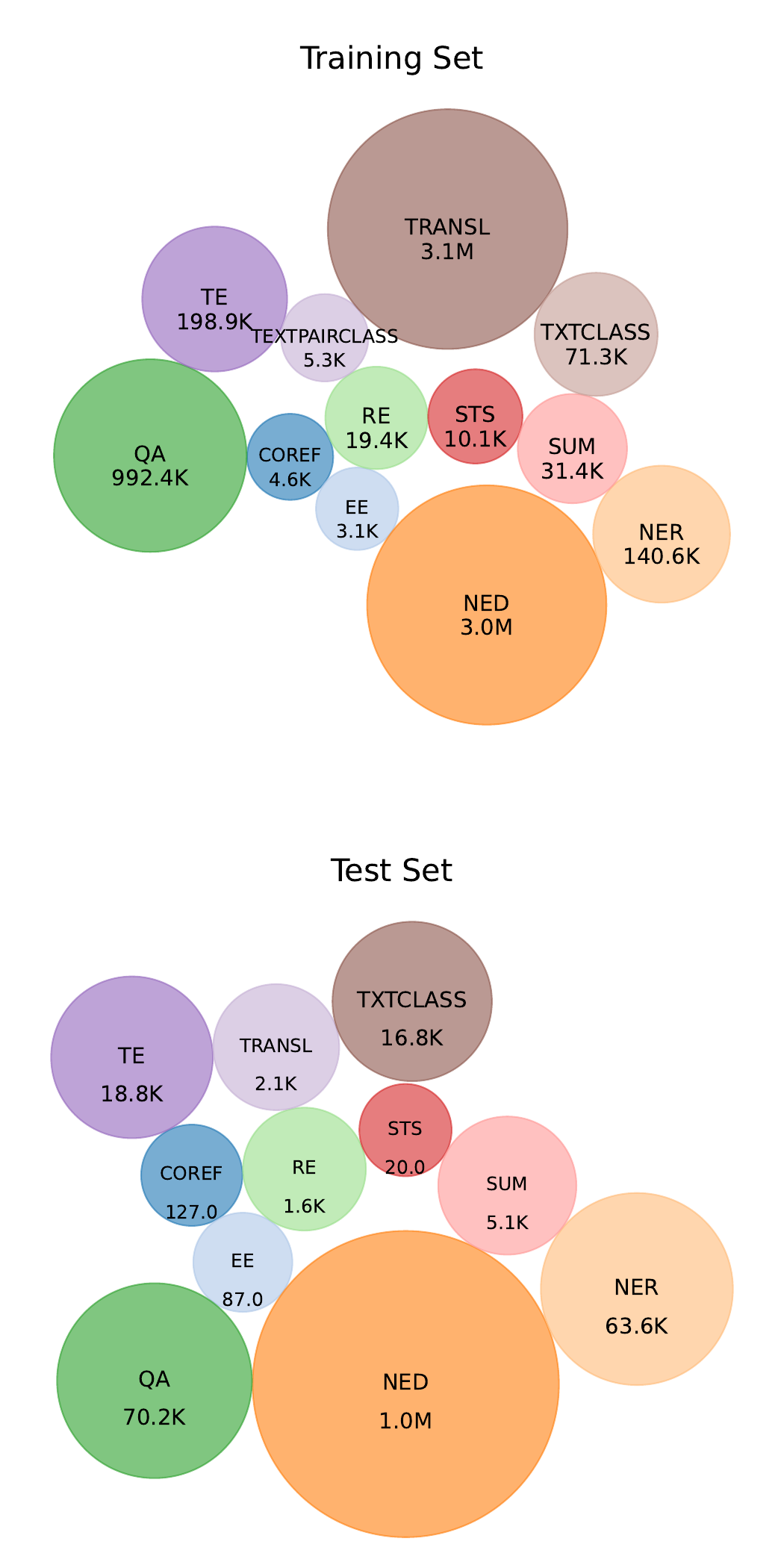}
    }
    \vspace{0pt}
    \hspace{0pt}
    \caption{\textsc{MedINST} overview.}
    \label{fig:overview}
\end{figure*}
\begin{figure}[h]
    \centering
    \includegraphics[width=0.48\textwidth]{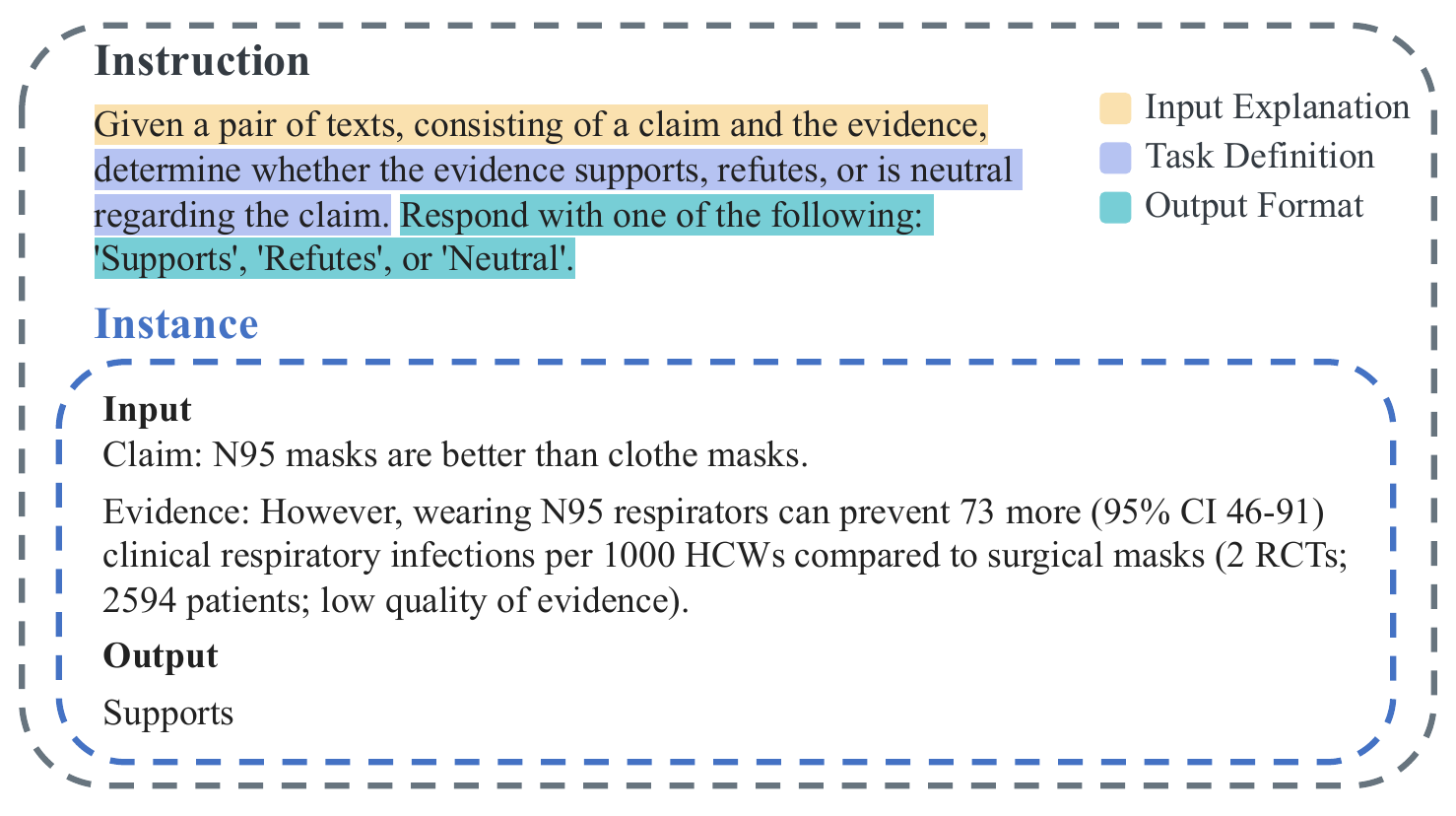}
    \caption{Instruction and instance example.}
    \label{fig:example}
\end{figure}

\begin{table*}[h]
\centering
\small
\resizebox{\textwidth}{!}{%
\begin{tabular}{cccccccccccccccc}
\toprule
 & & & \textbf{NER} & \textbf{RE} & \textbf{NED} & \textbf{QA} & \textbf{COREF} & \textbf{EE} & \textbf{TE} & \textbf{STS} & \textbf{TXTCLASS} & \textbf{TRANSL} & \textbf{SUM} & \textbf{TEXTPAIRCLASS} &\textbf{ALL}\\
\midrule
\large\multirow{6}{*}{\rotatebox[origin=c]{90}{\makecell[c]{\cellcolor{black!20}\textbf{\# Dataset}}}} &
\multirow{3}{*}{\textsc{MedINST}}   &\textbf{Train} &56 & 24 & 21 & 13 & 13 & 10 & 8 & 7 & 5 & 3 & 2 & 1 & 163\\
                                 &   &\textbf{Dev}   &30 & 11 & 10 & 8 & 10 & 7 & 5 & 1 & 4 & 1 & 1 & - & 88\\
                                 &   &\textbf{Test}  &37 & 9 & 12 & 10 & 2 & 1 & 8 & 1 & 5 & 1 & 1 & - & 87\\
\noalign{\vskip 6pt}\cline{2-16}\noalign{\vskip 6pt}

& \multirow{3}{*}{\textsc{MedINST32}}   &\textbf{Train}  &43 & 21 & 19 & 10 & 11 & 9 & 5 & 6 & 3 & 2 & 1 & 1 & 131\\
                                 &   &\textbf{Dev}       &19 & 9 & 9 & 6 & 8 & 6 & 5 & - & 2 & - & - & - & 64\\
                                 &   &\textbf{Test}      &13 & 3 & 2 & 3 & 2 & 1 & 3 & 1 & 2 & 1 & 1 & - & 32\\
\midrule
\noalign{\vskip 3pt}
\multicolumn{3}{c}{\large\cellcolor{black!20}\textbf{\# Instruction/Task}}        &49 & 23 & 19 & 9 & 7 & 9 & 3 & 3 & 5 & 3 & 2 & 1 & 133 \\
\noalign{\vskip 3pt}
\bottomrule
\end{tabular}}
\setlength{\abovecaptionskip}{5pt}
\caption{Dataset statistics across various categories.}
\label{tab:dataset_statistics}
\end{table*}

We curate \textsc{MedINST} by collecting 98 well-adopted biomedical datasets from 12 task categories and reformulating them into 133 tasks. All tasks are regarded as text generation task and the data are formatted to instruction-following samples. The instructions are human annotated and tailored for each dataset/task. Figure \ref{fig:overview} (a) depicts a visualization of the dataset composition of \textsc{MedINST}. 

\subsection{Tasks}
Figure \ref{fig:overview} (b) shows the number of samples included in each task categories. We adopted 12 categories of tasks, where each may have several sub-categories. The categories are as follows:
\paragraph{Named Entity Recognition (NER)} NER is a task in natural language processing that involves identifying and classifying key information entities. In the biomedical field, NER involves detecting and extract key entities such as diseases, drugs, genes, and other relevant biological terms within biomedical texts. In \textsc{MedINST}, 56 NER datasets are collected, including the most commonly used BC5CDR \cite{liBioCreativeCDRTask2016}, JNLPBA \cite{collierIntroductionBioentityRecognition2004}, LINNAEUS \cite{gernerLINNAEUSSpeciesName2010}, etc. We have created a unified instruction template for the NER task and made variations based on the specific requirements of each dataset. We have divided the NER task into two sub-categories, differing in output format. Sub-category 1 requires labeling each word in the input text using the BIO format, while Sub-category 2 requires directly outputting all detected entities that meet the criteria. In Sub-category 1, the input for each instance is a single sentence, whereas in Sub-category 2, the input is an entire passage. This adds diversity to the NER task and creates different levels of difficulty, thereby enhancing the model's stability in handling various output requirements and understanding longer texts.

\paragraph{Named Entity Disambiguation (NED)} The NED task involves determining the correct identity of named entities in a text by linking them to a specific entry in a knowledge base. Most of the NER datasets contain annotations for entity disambiguation. The NED task has also been repurposed into two difficulty levels. The AskAPatient and TwADR datasets \cite{limsopathamNormalisingMedicalConcepts2016} are used to create simpler tasks, where the input includes a specified biomedical entity and its context, and the requirement is to output its identifiers in the corresponding database. Other dataset such as BioRelEx \cite{khachatrianBioRelExBiologicalRelation2019}, CPI \cite{doringAutomatedRecognitionFunctional2020}, MedMentions \cite{mohanMedMentionsLargeBiomedical2019}, etc. have been reformatted into more challenging tasks, requiring the extraction of relevant biomedical entities from the given text and providing the corresponding identifiers for each entity. In additional, MeDAL dataset \cite{wenMeDALMedicalAbbreviation2020} has also been included in the NED task, which is a medical text dataset curated for abbreviation disambiguation. We include a total of 23 datasets in the NED task category.

\paragraph{Relation Extraction (RE)} RE involves identifying and categorizing the relationships between entities within a given text. We utilize 24 datasets for RE task, including AnEM \cite{ohtaOpendomainAnatomicalEntity2012}, BioNLP 2011 REL \cite{pyysaloOverviewEntityRelations2011}, etc. We simplified the task by listing all the possible relation types in the instruction for each dataset. The language model is prompted to extract all possible triples from the input text. 

\paragraph{Coreference Resolution (COREF)} COREF is the task of determining which words or phrases in a text refer to the same entity. We used 13 datasets for this task category, most of which come from the BioNLP Shared Task. In addition, the MLEE \cite{pyysaloEventExtractionMultiple2012} and PDR \cite{kimCorpusPlantDisease2019} datasets have also been included.

\paragraph{Question-Answering (QA)} Multiple types of QA are collected, including yes/no, yes/no/maybe, factoid, multi-choice, etc. In this category, 10 datasets are employed and reformatted. For multiple-choice QA, we write out the full options in the output rather than assigning letters or numbers to each option.

\paragraph{Textual Entailment (TE)} Determining whether two texts contradict each other and whether a statement aligns with the facts is crucial in the medical field. In this category, we re-format 6 fact-checking datasets, FEVER \cite{thorneFEVERLargescaleDataset2018}, HealthVer \cite{sarroutiEvidencebasedFactCheckingHealthrelated2021}, SciFact \cite{waddenFactFictionVerifying2020}, PubHealth \cite{kotonyaExplainableAutomatedFactChecking2020}, etc., into claim-evidence pairs. These datasets range from the general scientific domain to specific medical domains, such as COVID-19. Moreover, MEDIQA-RQE \cite{benabachaOverviewMEDIQA20192019} is incorporated as a question entailment task, i.e. determine whether the meaning of one question can be inferred from another question. As a classic task in the TE category, the premise-hypothesis entailment task is represented by the SciTail dataset \cite{khotSciTaiLTextualEntailment2018}.

\paragraph{Text Classification (TXTCLASS)} 
The text classification task involves assigning predefined categories or labels to a given piece of text based on its content. Although the input and output formats for text classification tasks are relatively fixed, the definitions and objectives of each task are highly diverse. Therefore, it is challenging to use a template to standardize this type of instruction. To ensure the quality of the instructions, each task within this category is entirely crafted manually. We collect 5 datasets, SciCite \cite{cohanStructuralScaffoldsCitation2019}, Hallmarks-of-Cancer \cite{bakerAutomaticSemanticClassification2016}, BC7-LitCovid \cite{chenMultilabelClassificationBiomedical2022}, MedDialog \cite{zengMedDialogLargescaleMedical2020} and GEOKhoj-v1\footnote{\url{https://github.com/ElucidataInc/GEOKhoj-datasets/tree/main/geokhoj_v1}}, for this category.

\paragraph{Semantic Similarity (STS)} The Semantic Similarity task aims at measuring how similar the meanings of two pieces of text are to each other. Originally a regression task with similarity scores as outputs, we have redefined it as a classification task by categorizing the similarity scores of all datasets into six integer levels from 0 to 5, where 0 indicates completely unrelated and 5 indicates highly similar. This task category includes 7 datasets, e.g. Bio-SimVerb, Bio-SimLex \cite{chiuBioSimVerbBioSimLexWidecoverage2018} BIOSSES \cite{soganciogluBIOSSESSemanticSentence2017}, MQP \cite{mccreeryEffectiveTransferLearning2020}, etc.

\paragraph{Event Extraction (EE)} EE task requires identifying and categorizing events, such as biological processes or interactions, within biomedical texts. The Event Extraction (EE) task is typically complex, with events in documents often containing nested structures. To format the EE task as a text generation task, we simplify it according to the BioNLP 2009 Core Event Detection subtask \cite{kimOverviewBioNLP092009}. We only detect events within a given range of types and their primary arguments. Note that primary arguments must be a biomedical entity within the text; we do not consider cases where primary arguments refer to another event.

\paragraph{Translation (TRANSL)} We have included the MuchMore \cite{buitelaarMultiLayeredXMLBasedApproach2003}, ParaMed \cite{liuParaMedParallelCorpus2021} and SciELO \cite{soaresLargeParallelCorpus2018} datasets translated from German, Chinese, and Spanish into English.

\paragraph{Text Pair Classification (TEXTPAIRCLASS)} For this category, we employ a sentiment analysis dataset, the Medical-Data\footnote{\url{https://www.kaggle.com/datasets/arbazkhan971/analyticvidhyadatasetsentiment}} 
, which analyzing the sentiment in a text where a drug is mentioned to determine whether the sentiment towards the drug is positive, negative, or neutral.

\paragraph{Summarization (SUM)} Summarization is also crucial for the application of LLMs in the biomedical field. In this category, we use the MeQSum \cite{benabachaSummarizationConsumerHealth2019} and Multi-XScience \cite{luMultiXScienceLargescaleDataset2020}  datasets. MeQSum presents patient questions, often in the form of lengthy texts, and the task requires capturing the main concern of these questions and providing a concise rewrite. Multi-XScience is a multi-document summarization task, which requires generating a related work section for a given article based on its abstract and the abstracts of some cited references.

The complete dataset collection details are listed in the Appendix \ref{sec:data_collection}.

\subsection{Instruction Construction}
All instructions are written according to a unified schema to ensure their quality. An instruction includes the following elements:
\paragraph{Input Explanation} 
The instruction first specifies the structure of the input. For example, for NER, the given input is typically a sentence or a passage; for QA tasks, the input can be a question alone, a question with context, or a question with context and options. We describe the elements included in the input for each dataset's task individually, avoiding the use of generalized descriptions.
\paragraph{Task Definition} The instructions include an explanation of the task and the specific actions the model needs to perform. The task definition is tailored to the content of each dataset and specifies any optional parameters. For example, the definition for the SciCite \cite{cohanStructuralScaffoldsCitation2019} task is "Classify the intent of the citation within this context. Intents are: [background, method, result]." avoiding vague instructions like "Classify the text into [background, method, result]."

\paragraph{Output Format} 
Here we specify the format of the output. In MedINST, we adopt formats corresponding to the complexity of the output content. For open text generation, the output is generally plain text; for classification tasks, multiple labels are separated by commas; for tasks like NER, where the output biomedical entities may contain various special characters, we enclose them in square brackets. For complex outputs, the instruction will provide a template example of the output format. 

After drafting instructions according to the abovementioned elements, we further proofread them to make them more concise and aligned with natural human instructions, avoiding rigid, structured descriptions. Appendix \ref{sec:ins_example} presents the examples of instructions.

\subsection{\textsc{MedINST32} Benchmark Construction}
\label{sec:medinst32}

Using the \textsc{MedINST} as a meta dataset, we carefully curate \textsc{MedINST32}, a challenging benchmark that covers 32 tasks with different difficulties to evaluate LLMs' performance across various medical-related tasks comprehensively. Unlike previous works, the tasks selected for \textsc{MedINST32} encompass different difficulty levels, including \textit{knowledge difficulty} and \textit{instruction difficulty}. 
Specifically, knowledge difficulty assesses the model's amount of biomedical knowledge, such as understanding levels of biomedical terms and their relationships, while instruction difficulty evaluates the model's understanding and adherence to instructions.
We divide difficulty into four categories and choose tasks from simplest (e.g., acronym completion) to hardest (e.g., RE, EE). Moreover, two positive examples are offered for each tasks.
See more details in Appendix \ref{sec:benchmark_detail}.

\section{Experiments}

\subsection{Setup}

\paragraph{Problem Formulation.}
We combine the training sets to train multi-task biomedical models. Given an instruction $Inst_t$ for a task $t$, and the dataset $(X_t, Y_t)$, multi-task models learns a map $M_t: (Inst_t, x)\rightarrow y$, where $(x, y)\in(X_t, Y_t)$. After learning a set of maps ${M_1, M_2, ..., M_T}$, the multi-task models can generalize to unseen tasks $i \in \{T+1, T+2, ..., T+N\}$ and approximate the maps $M_i$, where $M_i: (Inst_i, x)\rightarrow y$, $(x,y)\in(X_i, Y_i)$.

\paragraph{Training Data.}

Our goal is to test the generalization ability of LLMs on unseen tasks after instruction tuning multiple biomedical tasks. Once we have selected the 32 tasks in \textsc{MedINST32} (Sec. \ref{sec:medinst32}), we use the training set of the \textit{remaining} tasks from \textsc{MedINST} for multi-task fine-tuning. 
Since the \textsc{MedINST} training set is too large and a large number of training instances per task do not help generalization in instruction finetuning \citep{wangSuperNaturalInstructionsGeneralizationDeclarative2022}, we sample 100K samples to train our multi-task biomedical LLMs, denoted as \textbf{MI32}.
We select an equal number of samples from each task category to ensure balance across all tasks.

\paragraph{Evaluation setup.}

Following \citet{wangSuperNaturalInstructionsGeneralizationDeclarative2022}, we limit the test set for large size datasets aiming at efficient evaluation.
We observe that models not fine-tuned on MedINST sometimes struggled to output according to the instructions, posing challenges for post-processing and metric calculation. To ensure a fair comparison, we use few-shot prompts for baseline models during evaluation. Each test task is provided with two examples to help zero-shot models output in the standard format. Appendix \ref{sec:imp_detail} details the implementation of training and evaluation.

\paragraph{Model.} 
We fine-tune the instruction-tuned LLaMA-3 (8B; \citealp{llama3_2024}) and MMed-LLaMA-3 (8B; \citealp{qiu2024building}) on the aforementioned \textbf{MI32} training set and derive \textbf{LLaMA3-MI32} and \textbf{MMedL3-MI32}, respectively.
Additionally, we fine-tune LLaMA-3 on the 100K samples from \textsc{MedINST}, where the training sets of the datasets in \textbf{MI32} are exposed to the model, to produce \textbf{LLaMA3-MI}, as an oracle model.

\paragraph{Baselines.}
As a direct comparison, we compare our \textbf{LLaMA3-MI32} fine-tuned on \textbf{MI32} with its base version, \textbf{LLaMA3}. 
Since MMed-LLaMA-3 is a foundation model that has not been instruction fine-tuned, to make a fair comparison, we use MMed-LLaMA-3-EnIns \citep{qiu2024building}, which is fine-tuned on the English medical instruction dataset from PMC-LLaMA \cite{wuPMCLLaMABuildingOpensource2023}.
We denote it as \textbf{MMedL3-EnIns}.
In addition, we compare \textbf{BioMistral}, an open-source LLM further pretrained on PubMed Central utilizing the instruction fine-tuned version of Mistral-7B \cite{jiangMistral7B2023} and \textbf{GPT-4o}, an advanced variant of GPT-4, excels in the biomedical domain with enhanced capabilities for understanding and generating complex medical and scientific texts. 

\paragraph{Metrics.}
Inspired by BLURB \cite{guDomainSpecificLanguageModel2021}, we select appropriate metrics for each task in \textsc{MedINST32}, including \textbf{Rouge-L}, \textbf{Entity F1} (Entity-level F1), \textbf{Label F1} (Label-level F1), \textbf{MSE} (Mean Squared Error) and \textbf{EM} (Exact Match).
Entity-level F1  measures the overlap between the entities detected by the models and the ground truth, which is calculated by each data sample. Label-level F1 is calculated from the entire dataset to measure the similarity between the model's predictions and the labels.

\subsection{Results}
Table \ref{tab:result_normal} presents the evaluation results of our models and baselines on \textsc{MedINST32}. 
As an oracle model, \textbf{MMedL3-MI} demonstrated excellent performance across various difficulty levels, outperforming \textbf{GPT-4o} in 25 tasks. This highlights the significant impact of the \textsc{MedINST} dataset in enhancing the overall performance of models on biomedical tasks. 

The two zero-shot models, \textbf{LLaMA3-MI32} and \textbf{MMedL3-MI32}, showed significant generalization improvements over their base models in most unseen tasks. They respectively outperformed GPT-4o in 15 and 13 tasks. However, surprisingly, \textbf{MMedL3-MI32}, which used MMed-LLaMA-3 (further pretrained on biomedical corpora) as its base model, lagged behind LLaMA3-MI32 in 22 tasks. This indicates that using further pretraining to specialize a general LLM to the biomedical domain may not be as effective as instruction fine-tuning, especially considering the substantial computational resources required for pretraining. This also underscores the necessity of building a comprehensive biomedical instruction meta-dataset.

MMedL3-EnIns was fine-tuned on 500K medical question-answering data, which includes training data from MedQA and PubMedQA that appeared in \textsc{MedINST32}. Despite using few-shot prompting, its performance on \textsc{MedINST32} was still unsatisfactory. It even significantly lagged behind in QA tasks, especially in MedQA, achieving only a 15.40 accuracy. This highlights the necessity of reformulating tasks to improve model generalization capabilities: training models to output in a single format alone increases the risk of overfitting.

\begin{table*}[ht!]
\centering
\resizebox{\textwidth}{!}{%
\begin{tabular}{ccccccccccc}
\toprule
\multirow{3}{*}{\textbf{Category}} & \multirow{3}{*}{\textbf{Dataset}} &\multirow{3}{*}{\textbf{Difficulty Level}} & \multirow{3}{*}{\textbf{Metric}} & \multicolumn{7}{c}{\textbf{Model}} \\ 
\cmidrule(lr){5-11}
 & & & &\textbf{LLaMA3} & \textbf{BioMistral}  & \textbf{MMEDL3-EnIns} & \textbf{GPT-4o} &\textbf{LLaMA3-MI32} & \textbf{MMEDL3-MI32} & $\textbf{LLaMA3-MI}^{\dagger}$ \\ 
 \cmidrule(lr){5-8}\cmidrule(lr){9-11}
 &&&&\multicolumn{4}{c}{\textbf{(Few Shot)}}&\multicolumn{3}{c}{\textbf{Ours (Zero Shot)}} \\
 \cmidrule(lr){1-8} \cmidrule(lr){9-11}
\multirow{13}{*}{\textbf{NER}} & NCBI-disease                      &2   & Label-F1   & 51.67 & 24.00 & 30.59 & 47.57 & \underline{78.55} & 78.20 & \textbf{84.61}\\ 
                               & BC5CDR                            &2   & Label-F1   & 58.68 & 33.86 & 28.77 & 75.11 & \underline{81.28} & 73.57 & \textbf{87.39}\\  
                               & AnEM                              &3   & Entity-F1  & 8.20 & 3.66 & 1.72 & \underline{37.44} & 32.03 & 31.38 & \textbf{49.44}\\ 
                               & BioNLP-2009                       &2   & Entity-F1  & 30.33 & 22.24 & 19.71 & 57.83 &  76.06 & \underline{78.61} & \textbf{80.74}\\ 
                               & BioNLP-2011-GE                    &2   & Entity-F1  & 29.60 & 20.97 & 14.40 & 57.43 & 76.29 & \underline{79.89} & \textbf{80.39}\\ 
                               & BioNLP-2011-ID                    &3   & Entity-F1  & 32.83 & 18.45 & 21.19 & \textbf{68.59} & 51.80 & 50.80 & \textbf{76.26}\\ 
                               & BioNLP-2011-REL                   &2   & Entity-F1  & 30.14 & 22.73 & 20.26 & 59.01 & 75.93 & \underline{78.66} & \textbf{80.41}\\ 
                               & BioNLP-2013-CG                    &3   & Entity-F1  & 24.46 & 10.49 & 8.63 & \underline{57.59} & 56.36 & 51.61 & \textbf{72.32}\\ 
                               & BioNLP-2013-GE                    &2   & Entity-F1  & 16.98 & 15.49 & 13.29 & 43.74 & \textbf{71.59} & 71.25 & \underline{71.32}\\ 
                               & BioNLP-2013-GRO                   &4   & Entity-F1  & 10.34 & 4.14 & 2.91 & \textbf{37.79} & 12.48 & 12.86 & \underline{35.13}\\ 
                               & BioNLP-2013-PC                    &3   & Entity-F1  & 31.96 & 19.45 & 19.57 & \underline{68.75} & 62.94 & 61.38 & \textbf{82.05}\\ 
                               & BioRED                            &3   & Entity-F1  & 29.38 & 16.45 & 16.33 & 60.73 & \underline{74.01} & 72.45 & \textbf{78.76}\\ 
                               & tmVar-v3                          &3   & Entity-F1  & 16.34 & 8.96 & 0.39 & 42.08 & \underline{58.46} & 56.77 & \textbf{63.22}\\  \midrule
\multirow{3}{*}{\textbf{QA}}   & BioASQ-Task-B-yesno               &1   & Label-F1   & 91.62 & 67.57 & 91.82 & \underline{93.52} & 93.10 & 86.19 & \textbf{93.87}\\ 
                               & PubMedQA-labeled                  &2   & Label-F1   & 50.85 & 23.73 & 48.28 & \underline{56.11} & 53.81 & 53.65 & \textbf{59.94}\\ 
                               & MedQA                             &2   & EM         & 49.25 & 24.51 & 15.40 & \textbf{81.93} & 47.68 & 45.72 & \underline{53.26}\\  \midrule
\multirow{3}{*}{\textbf{TE}}   & SciFact                           &2   & Label-F1   & 42.09 & 36.33 & 33.69 & \underline{92.61} & 85.85 & 84.14 & \textbf{95.06}\\  
                               & ManConCorpus                      &2   & Label-F1   & 66.66 & 29.92 & 51.83 & 60.09 & 68.20 & \underline{68.57} & \textbf{69.14}\\ 
                               & CoVERt                            &1   & Label-F1   & 82.24 & 47.77 & 55.87 & \underline{93.76} & 91.15 & 93.49 & \textbf{96.93}\\   \midrule
\multirow{2}{*}{\textbf{TXTCLASS}} & Hallmarks-of-Cancer           &2   & Entity-F1  & 45.33 & \textbf{54.77} & 11.93& 42.40 & 44.01 & 32.65 & \underline{45.84}\\ 
                                   & MedDialog                     &1   & Label-F1   & 91.34 & 86.02 & 56.52 & \underline{98.77} & 96.72 & 77.67 & \textbf{100.00}\\  \midrule
\multirow{2}{*}{\textbf{NED}}   & MeDAL                            &2   & EM         & 21.6 & 15.90 & 17.00 & \textbf{59.40} & 28.90 & 30.00 & \underline{36.60}\\ 
                               & tmVar-v3-NED                      &4   & Entity-F1  & 0.18 & 0.05 & 0.00 & \textbf{7.45}  & \underline{2.84} & 0.78 & 1.10\\   \midrule
\multirow{3}{*}{\textbf{RE}}   & AnEM-RE                           &4   & Entity-F1  & 2.56 & 0.00 & 5.13 & \textbf{25.64} & 0.20 & 1.54 & \underline{16.24}\\  
                               & BC5CDR-RE                         &4   & Entity-F1  & 4.28 & 6.27 & 3.34 & 9.46 & \underline{14.21} & 13.69 & \textbf{27.93}\\  
                               & BioInfer-RE                       &4   & Entity-F1  & 18.74 & 9.73 & 8.86 & 17.49 & \underline{28.06} & 26.23 & \textbf{32.83}\\   \midrule
\multirow{2}{*}{\textbf{COREF}} & AnEM-COREF                       &1   & Entity-F1  & 34.52 & 14.29 & 21.43 & \underline{82.20} & \textbf{100.00} & \textbf{100.00} & \textbf{100.00}\\  
                                & MLEE-COREF                       &1   & Entity-F1  & 54.17 & 26.55 & 25.66 & 79.97 & \textbf{99.12} & \underline{98.23} & 95.72\\   \midrule
\multirow{1}{*}{\textbf{SUM}}  & Multi-XScience                    &2   & Rouge-L    & \underline{13.28} & 11.61 & 10.36 & 12.78 & 11.61 & 11.57 & \textbf{14.51}\\  \midrule
\multirow{1}{*}{\textbf{EE}}   & MLEE-EE                           &4   & Entity-F1  & 0.96 & 0.19 & 0.09 & 9.88 & \textbf{30.47} & \underline{28.61} & 27.48\\   \midrule
\multirow{1}{*}{\textbf{STS}}  & BIOSSES                           &1   & MSE$\downarrow$  & 2.05 & 4.15 & 4.15 & \textbf{0.6} & \underline{1.05} & 2.15 & 1.20\\   \midrule
\multirow{1}{*}{\textbf{TRANSL}} & ParaMed                         &2   & Rouge-L    & 47.51 & 50.49 & 46.49 & \textbf{63.08} & 49.01 & 49.65 & \underline{59.32}\\ 
 \bottomrule
\end{tabular}%
}
\small\caption{
\footnotesize{
Test results of various models on \textsc{MedINST32}. $\dagger$ indicates that the training sets of LLaMA3-MI includes the corresponding training sets of the datasets used by \textsc{MedINST32}, whereas other models have not seen the \textsc{MedINST32} dataset. $\downarrow$ represents that a lower score is better, while for other metrics, a higher score is better. The best and second-best results for each row are highlighted in bold and underlined, respectively. For the baselines, we use a few-shot prompt, providing two examples in the instruction. For the fine-tuned models, we use a zero-shot prompt.
}
}
\label{tab:result_normal}
\end{table*}

\begin{table}[ht]
\centering
\resizebox{\columnwidth}{!}{%
\begin{tabular}{cccccccc}
\toprule
\multirow{2}{*}{\textbf{Method}} & \multicolumn{6}{c}{\textbf{MMLU}} & \\
\cmidrule{2-7}
& An & CK & CB & CM & MG & PM & Avg.  \\
\midrule
\textbf{BioMistral}      &48.89  &66.42  &63.19  &58.38  &70.00  &58.46  &60.88\\
\textbf{MMedL3}    &65.19          &70.19          &72.22          &55.49                  &74.00                  &66.91          &67.03  \\
\textbf{MMedL3-EnIns}     &68.15  &64.91  &71.52  &59.53  &76.00  &72.79  &68.32\\
\textbf{LLaMA3}         &67.41          &\textbf{76.60} &\textbf{80.56} & \textbf{67.63}        &82.00                  & 72.06         & 73.92 \\
\textbf{MMedL3-MI} (Ours)     &64.44          &67.92          &71.53          &58.96                  &74.00                  & 66.54         &66.76   \\
\textbf{LLaMA3-MI} (Ours)      &\textbf{68.15} &75.47          &75.00          &\textbf{67.63}         &\textbf{83.00 }        &\textbf{77.21} &\textbf{74.38}  \\

\bottomrule
\end{tabular}
}
\caption{
\footnotesize{
Multiple-choice accuracy evaluation on MMLU-Medicine, a subset of MMLU benchmark. The subjects used are anatomy (An), clinical knowledge (CK), college biology (CB), college medicine (CM), medical genetics (MG) and professional medicine (PM).
}
}
\label{tab:mmlu}
\end{table}

\begin{figure}[ht]
    \centering
    \subfigure[Performance with varying training data sizes.]{
        \includegraphics[width=0.35\textwidth]{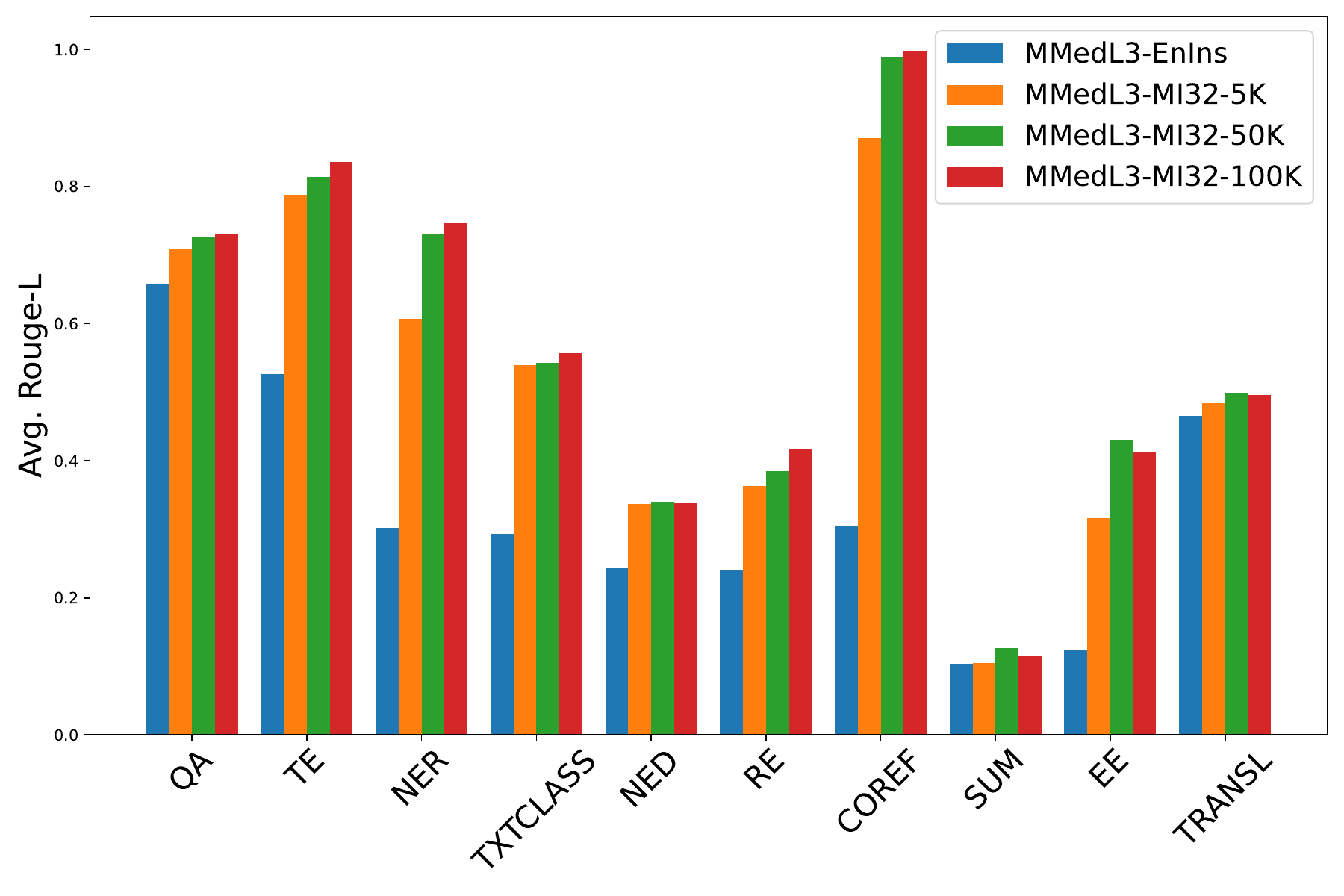}
    }
    \hfill
    \subfigure[Performance with varying model parameter sizes.]{
        \includegraphics[width=0.35\textwidth]{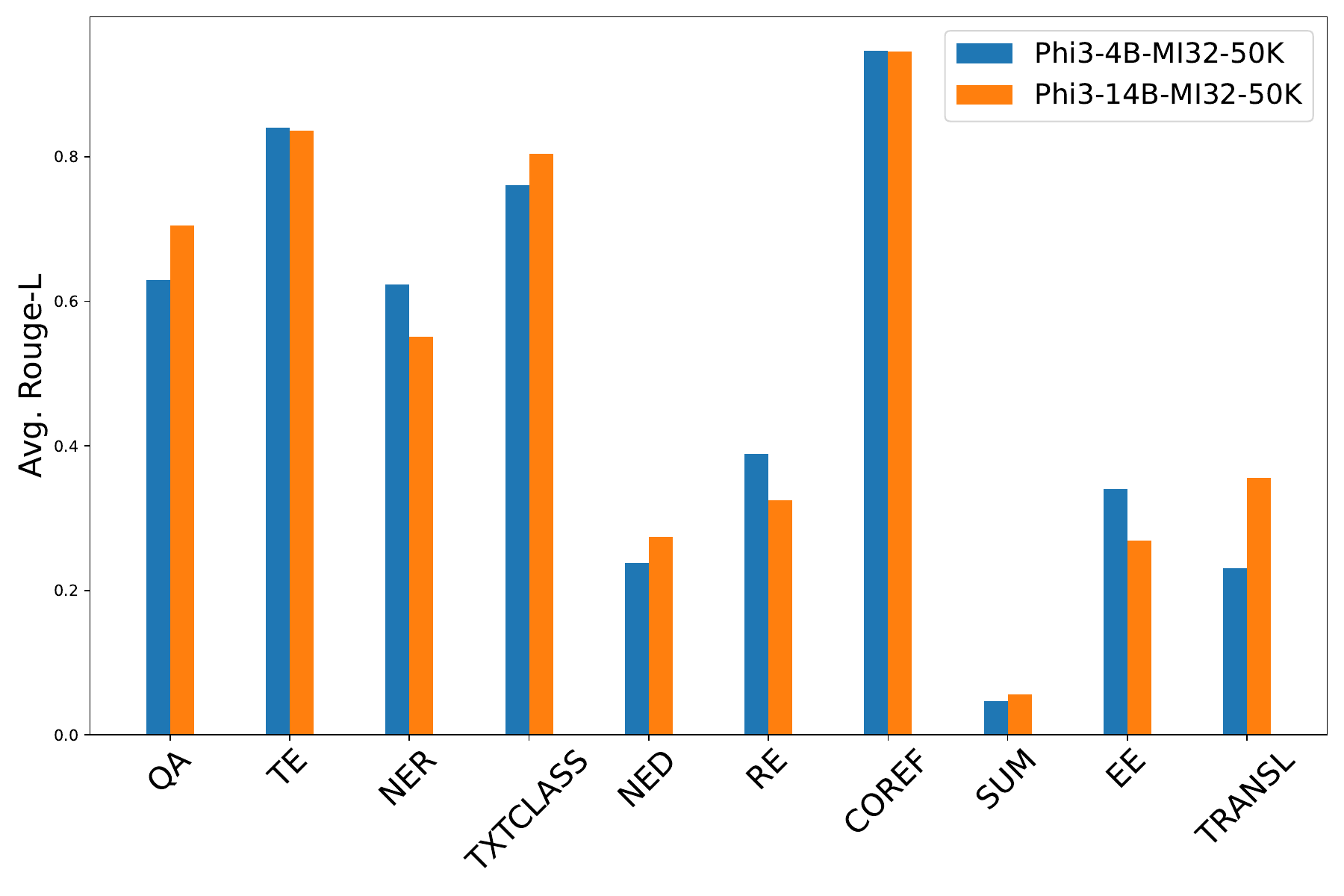}
    }
    \caption{\footnotesize{Training sample and model parameter scale analysis.}}\label{fig:trend}
\end{figure}

\subsection{Ablation  Analysis}
We design experiments to explore the impact of the number of training samples and model parameters on finetuning performance. We employ the same strategy to sample 5K and 50K instances from the \textsc{MI32} training set for training two additional MMedL3-MI32 models for comparison. Additionally, we trained both 4B and 14B versions of Phi-3 using the 50K dataset.

In Figure \ref{fig:trend}, we calculate the average Rouge-L score for each task category to measure the performance of the models. In (a), it can be seen that as the number of training samples increases, the model's overall performance improves. However, performance deteriorates with increased sample size in tasks such as summarization (SUM) and event extraction (EE). This is because as sampling expands, the proportion of smaller datasets decreases, leading to data imbalance, which causes uneven learning progress across different tasks. Part (b) demonstrates unexpected results regarding the scale of model parameters. Phi-3-14B performs less than the 4B version in three core tasks for the biomedical field: NER, RE, and EE. A possible reason is that larger models require more data to be fully optimized and achieve generalization performance on unseen biomedical tasks. Specialized tasks such as Named Entity Recognition (NER), Relation Extraction (RE), and Event Extraction (EE) in the biomedical field are more susceptible to overfitting on small data sets compared to more general tasks like summarization (SUM).

\subsection{Evaluation on Public English Benchmarks}
The Massive Multitask Language Understanding (MMLU; \citealp{hendrycks2021measuring}) is a benchmark that evaluates language models across various QA tasks and subjects. We train \textbf{MMedL3-MI} using the same 100K dataset that was used to train \textbf{LLaMA3-MI}. The models are tested on 6 medical-related subtasks of MMLU. Table \ref{tab:mmlu} exhibits the result. As seen, \textbf{LLaMA3-MI} and \textbf{MMedL3-MI} perform similarly to the baseline model on MMLU-Medicine. Additionally, note that \textbf{LLaMA3-MI} and \textbf{MMedL3-MI} are multitask models in the biomedical field, capable of handling various other, more challenging biomedical tasks.

\section{Conclusion}

In this paper, we introduce an instruction meta-dataset \textsc{MedINST} comprising 133 biomedical tasks across 12 task categories and a challenging benchmark \textsc{MedINST32} for evaluating multitask biomedical models. 
Through various experiments, we train multiple biomedical models and demonstrate their strong generalization performance on biomedical tasks using our dataset. Due to resource constraints, we trained only on a small subset and 8B models. Using the full dataset and larger models may lead to further improvements, which is left for future work. Our work lays the foundation for developing better-performing biomedical LLMs.

\section*{Limitations}

We identify our limitations as follows. 

First, due to computational resource constraints, we conducted our experiments with limited data and model sizes. We used the LoRA technique to finetune our model, which might limit the learning outcomes. Full-parameter finetuning could potentially yield better results. In future work, we will continue to explore ways to further enhance the performance of LLMs on biomedical-related tasks. 

Currently, the MedINST dataset only includes single-turn dialogues, which may limit the model's ability to generalize to multi-turn dialogue tasks. Therefore, in the future, we plan to incorporate multi-turn instruction samples. 

Additionally, the current dataset is primarily in English, with other languages featured in the TRANSL tasks, so another direction for future work is to continue expanding the multilingual data.

\bibliography{custom}

\clearpage

\appendix
\section{Instruction Benchmark Construction}\label{sec:benchmark_detail}
To comprehensively evaluate the model's performance across various medical-related tasks, we selected 32 tasks from each task category in the \textsc{MedINST} dataset to establish a new biomedical benchmark, \textsc{MedINST32}. The tasks selected for benchmarking encompass different levels of difficulty. This includes two aspects: knowledge difficulty and instruction difficulty. Knowledge difficulty assesses the amount of biomedical knowledge the model possesses, such as understanding categories of biomedical terms and their relationships. For basic-level assessment, we chose tasks like acronym completion (MeDAL). Intermediate-level tasks include various NER, QA, TE, and TXTCLASS tasks. Finally, we included more challenging tasks like RE, EE, and tasks in NED that involve annotating identifiers. Instruction difficulty evaluates the model's understanding and adherence to instructions. This dimension was not considered in previous benchmark datasets. For example, in multichoice QA tasks, previous works often labeled each option as A, B, C, etc., and the model only needed to respond with the corresponding label. In our QA task construction, we require the model to output the selected option as it is, which increases the task difficulty and reduces the chance of the model bypassing with simple letter responses. Additionally, we construct different instructions for similar tasks. For instance, in NER tasks, we developed two types of instructions: one requiring the model to repeat each word in the text in a BIO format and label them one by one, and the other asking the model to directly extract all biomedical entity mentions and annotate their categories.

For each task in \textsc{MedINST32}, we provide two positive examples. For tasks that have a training set, we select two examples from their training set. If a task does not have a training set, we find the most similar task from all the test set tasks in \textsc{MedINST} and select from there. During selection, we strive to ensure that the two examples are diverse in content. For instance, in classification tasks, we choose examples with different labels. 

We remove all the datasets used in \textsc{MedINST32} from the \textsc{MedINST} training set to create the training set for \textsc{MedINST32}.

 We performed random sampling on a portion of tasks with abundant data resources to control the number of test data in each category to be roughly consistent. This helps to reduce the computational resource consumption for evaluation. The sample sizes are shown in Table \ref{tab:sampling_size}. For other datasets, we use the entire test set data.

\begin{table}[h!]
\centering
\begin{tabular}{lr}
\hline
\textbf{Dataset Name}      & \textbf{Sample Size} \\ \hline
NCBI-disease               & 100               \\ \hline
BC5CDR                     & 100               \\ \hline
BioNLP-2011-GE             & 100               \\ \hline
tmVar-v3                   & 100               \\ \hline
MeDAL                      & 1000              \\ \hline
ParaMed                    & 200               \\ \hline
Multi-XScience             & 200               \\ \hline
\end{tabular}
\caption{Sampling sizes for evaluation.}
\label{tab:sampling_size}
\end{table}

Overall, we provide a more comprehensive and challenging biomedical instruction benchmark compared to previous works.

\section{Instruction Examples}\label{sec:ins_example}
Table \ref{sec:ins_example} presents the instruction examples for each task categories. 
 Each instruction contains three parts: input explanation, task definition, and output format, which clearly tell the LLM how to complete the task. For each task within a category, the instruction can vary, thus requiring manual composition. However, for categories such as NED, RE, and EE tasks, the main body of the instruction is generic. We can efficiently edit the instruction by modifying some variable fields based on the metadata of each dataset, and these variable fields are highlighted in blue.
\begin{table*}[h]
    \centering
    \small
    \begin{tabularx}{\linewidth}{X}
    \toprule
    \textbf{QA} Given a question and context, select the correct answer from the provided options.\\
    \midrule
    \textbf{TE} Given a pair of texts, consisting of a claim and the evidence, determine whether the evidence supports, refutes, or is neutral regarding the claim. Respond with one of the following: `Supports', `Refutes', or `Neutral'.\\
    \midrule
    \textbf{NER} Given a sentence, label each disease, disease class and symptom entity using the BIO format. In BIO format, `B' indicates the beginning of an entity, `I' indicates the inside of an entity, and `O' indicates a token not part of any entity. Label each word in the format: `word [LABEL]'.\\
    \midrule
    \textbf{TXTCLASS} You are provided with a citation context. Classify the intent of the citation within this context. Intents are: \textcolor{blue}{[background, method, result]}.\\        \midrule
    \textbf{NED} You are provided with a text. Your objective is to identify and extract all chemical and disease entities mentioned in the text, maintaining the order in which they appear. For each entity, provide its corresponding database identifier from \textcolor{blue}{MESH}. The entities should be presented in the format: [entity1 <db\_name/db\_id>].\\       \midrule
    \textbf{RE} Given a text, identify and extract specified relations between anatomical entities mentioned within it. The specified relation types are \textcolor{blue}{[frag, Part-of]}. Relation explanation: frag: Frag relation marking coordination with ellipsis; Part-of: Part-of relation marking entity mention spanning a prepositional phrase. Present each relation in format as follows: [<entity1> <relation> <entity2>].\\     \midrule
    \textbf{COREF} Given a text and a specified \textcolor{blue}{anatomical entity}, identify and extract all co-references to that entity within the text. Present each co-reference entity in the following format: [co-reference entity].\\       \midrule
    \textbf{STS} Given two texts, evaluate their similarity and provide an integer score ranging from 0 to 5, where 0 indicates no similarity and 5 indicates high similarity.\\      \midrule
    \textbf{EE} Given a text, identify and extract the epecified types of bio-molecular events along with their primary arguments. The event type can be \textcolor{blue}{[Binding, Positive\_regulation, Phosphorylation, Regulation, Transcription, Localization, Gene\_expression, Protein\_catabolism, Negative\_regulation]}. Present each event in the format as follows: [<type> <trigger> <theme entity>].\\       \midrule
    \textbf{TRANSL} Translate the text from \textcolor{blue}{Chinese} to English.\\       \midrule
    \textbf{TEXTPAIRCLASS}  You are given a drug name and a piece of text. Analyze the sentiment in the text and determine whether the sentiment towards the drug is positive, negative, or neutral. Answer with `Positive', `Negative', or `Neutral'.\\      \midrule
    \textbf{SUM} Writing the related-work section of a paper based on its abstract and the articles it references.\\
    \bottomrule
    \end{tabularx}
    \caption{Instruction examples for each task category.}
    \label{tab:ins_example}
\end{table*}
\section{Implementation Details}\label{sec:imp_detail}
\paragraph{Training} For the baseline models, we used the LLaMA-3-8B-Instruct \footnote{\url{https://huggingface.co/unsloth/llama-3-8b-Instruct}} and MMed-LLaMA-3-8B \footnote{\url{https://huggingface.co/Henrychur/MMed-Llama-3-8B}} models available on Hugging Face. Due to limited computational resources, we employed Low-Rank Adaptation (LoRA) for parameter-efficient fine-tuning (PEFT). The LoRA rank was set to 8, targeting all linear layers, including $q\_proj$, $k\_proj$, $v\_proj$, $o\_proj$, $gate\_proj$, $up\_proj$, and $down\_proj$. The learning rate was set at 1.0e-4, with a batch size of 4 and gradient accumulation steps of 4. We used a cosine learning rate scheduler with a 0.1 ratio of warmup. For training, we ran 5 epochs with 5K data, and 3 epochs for 50K and 100K datasets. The training was conducted on a single 40GB A100 GPU.

\paragraph{Query Template} For the training and evaluation of all LLaMA-3 series models, we used the standard LLaMA-3 chat template. Table \ref{tab:llama3_template} shows an example. 
When constructing few-shot prompts, each example is treated as a round of dialogue and added before the query that needs an answer. Unlike the approach where instructions are only given in the first round of dialogue, we included instructions in each example. This is because for some tasks without a training set, we selected examples from the training sets of similar tasks, so the instructions in the examples may not completely match the instructions of the query. Table \ref{tab:query_example} demonstrates a query of a NER task.
\begin{table*}[h]
    \centering
    \begin{tabular}{p{0.8\textwidth}}
    \toprule
    <|begin\_of\_text|><|start\_header\_id|>\textcolor{red}{system}<|end\_header\_id|>\\
    \\
    \textcolor{red}{You are a helpful assistant.}<|eot\_id|><|start\_header\_id|>\textcolor{cyan}{user}<|end\_header\_id|>\\
    \\
    \textcolor{cyan}{Given an utterance, determine if it is from a doctor or a patient. Do i have covid 19?}<|eot\_id|><|start\_header\_id|>\textcolor{orange}{assistant}<|end\_header\_id|>\\
    \\
    \textcolor{orange}{patient}<|eot\_id|>\\
    \bottomrule
    \end{tabular}
    \caption{LLaMA-3 prompt template.}
    \label{tab:llama3_template}
\end{table*}

\begin{table*}[h]
    \centering
    \begin{tabularx}{\linewidth}{l|lX}
    \toprule
    \multirow{3}{*}{\textbf{Example 1}}&\textbf{Instrcution:}& You are provided with a text. Your objective is to identify, extract and classify all gene and protein entities mentioned in the text, maintaining the order in which they appear. Types are [Gene, DomainMotif, FamilyName]. The entities should be presented in the following format: [entity <type>].\\
    &\textbf{Input:}& Cloning, expression and localization of an RNA helicase gene from a human lymphoid cell
    ... ... cell line from a diffuse large B-cell lymphoma.\\
    &\textbf{Output:}& [RNA helicase <FamilyName>] [RNA helicase <FamilyName>] [p54 <Gene>] [RNA helicase <FamilyName>] [ME31B <Gene>] [ME31B <Gene>]\\
    \hdashline
    \multirow{3}{*}{\textbf{Example 2}}&\textbf{Instrcution:}& You are provided with a text. Your objective is to identify, extract and classify all gene variant entities mentioned in the text, maintaining the order in which they appear. Types are [DNAMutation, SNP, ProteinMutation]. The entities should be presented in the following format: [entity <type>].\\
    &\textbf{Input:}& A novel multidrug-resistance protein 2 gene mutation identifies a ... ... heterozygous mutation was significantly associated with the presence of pruritus.\\
    &\textbf{Output:}& [V1188E <ProteinMutation>]\\
    \hdashline
        \multirow{2}{*}{\textbf{Query}}&\textbf{Instrcution:}& You are provided with a text. Your objective is to identify, extract and classify all gene variant entities mentioned in the text, maintaining the order in which they appear. Types are [OtherMutation, Species, DNAAllele, DNAMutation, CellLine, SNP, ProteinMutation, ProteinAllele, Gene, AcidChange]. The entities should be presented in the following format: [entity <type>].\\
    &\textbf{Input:}& A novel single-nucleotide substitution, Glu 4 Lys ... ... Thus, our results suggest that Glu 4 Lys in the LTC4S might be associated with allergic diseases.\\
    
    \bottomrule
    \end{tabularx}
    \caption{Query example.}
    \label{tab:query_example}
\end{table*}
\newpage
\section{Extra Metrics for SUM and TRANSL Tasks}
We add additional metrics, BERT score and METEOR score, to evaluate the generated text on summarization and translation tasks. The evaluation results are presented in Table \ref{tab:sum_task} and Table \ref{tab:transl_task}.
\begin{table}[h!]
\centering
\resizebox{\linewidth}{!}{
\begin{tabular}{lcc}
\hline
\textbf{Model} & \textbf{BERTScore} & \textbf{METEOR Score} \\ \hline
LLaMA3 & 0.7467 & 0.1758 \\ \hline
BioMistral & 0.7253 & 0.1152 \\ \hline
MMEDL3-EnIns & 0.7314 & 0.1185 \\ \hline
GPT-4o & 0.8317 & 0.2333 \\ \hline
LLaMA3-MI32 (ours) & 0.7951 & 0.1566 \\ \hline
MMEDL3-MI32 (ours) & 0.7963 & 0.1220 \\ \hline
LLaMA3-MI (ours) & 0.8203 & 0.1592 \\ \hline
\end{tabular}}
\caption{SUM task: Multi-XScience results.}
\label{tab:sum_task}
\end{table}

\begin{table}[h!]
\centering
\resizebox{\linewidth}{!}{
\begin{tabular}{lcc}
\hline
\textbf{Model} & \textbf{BERTScore} & \textbf{METEOR Score} \\ \hline
LLaMA3 & 0.9000 & 0.3776 \\ \hline
BioMistral & 0.9101 & 0.3670 \\ \hline
MMEDL3-EnIns & 0.8888 & 0.3625 \\ \hline
GPT-4o & 0.9291 & 0.4661 \\ \hline
LLaMA3-MI32 (ours) & 0.9115 & 0.3933 \\ \hline
MMEDL3-MI32 (ours) & 0.9080 & 0.3781 \\ \hline
LLaMA3-MI (ours) & 0.9379 & 0.6126 \\ \hline
\end{tabular}}
\caption{TRANSL task: ParaMed results.}
\label{tab:transl_task}
\end{table}

\section{Dataset Collection}
Table \ref{tab:dataset_collection} lists all the dataset employed in \textsc{MedINST}. Because a single dataset might be reformulated into multiple tasks, we added suffixes to the names in the multi-task dataset. For example, BC5CDR appears in the NER, NED, and RE tasks. For the primary task, NER, we use the dataset's original name, and for the other two tasks, we append the respective suffixes to the dataset name.
\label{sec:data_collection}
\onecolumn
\begin{longtable}{lllll}
\caption{Dataset collection.} \\
\hline
Dataset & Task & Train & Dev & Test \\
\hline
\endfirsthead

\multicolumn{5}{c}%
{\tablename\ \thetable\ -- \textit{Continued from previous page}} \\
\hline
Dataset & Task & Train & Dev & Test \\
\hline
\endhead

\hline \multicolumn{5}{r}{\textit{Continued on next page}} \\
\endfoot

\hline
\endlastfoot

   BioASQ-Task-B-yesno &            QA &    15,568 &         0 &       813 \\
    BioASQ-Task-B-list &            QA &    11,687 &         0 &     1,000 \\
 BioASQ-Task-B-factoid &            QA &    16,389 &         0 &       724 \\
 BioASQ-Task-B-summary &            QA &    13,151 &         0 &       824 \\
   BiologyHowWhyCorpus &            QA &     1,269 &         0 &         0 \\
                BIOMRC &            QA &   700,000 &    50,000 &    62,707 \\
Evidence-Inference-2.0 &            QA &    10,056 &     1,233 &     1,222 \\
                 MedQA &            QA &    10,178 &     1,273 &     1,272 \\
                MedHop &            QA &     1,620 &       342 &         0 \\
             MEDIQA-QA &            QA &       312 &        25 &       150 \\
   PubMedQA-artificial &            QA &   200,000 &    11,269 &         0 \\
      PubMedQA-labeled &            QA &       450 &        50 &       500 \\
                  SciQ &            QA &    11,679 &     1,000 &     1,000 \\
                 FEVER &            TE &   145,449 &     9,999 &     9,999 \\
             HealthVer &            TE &    10,590 &     1,917 &     1,823 \\
             PubHealth &            TE &     9,804 &     1,214 &     1,233 \\
               SciFact &            TE &       868 &         0 &     1,189 \\
          ManConCorpus &            TE &         0 &         0 &     2,775 \\
                CoVERt &            TE &         0 &         0 &       212 \\
            MEDIQA-RQE &            TE &     8,588 &       302 &       230 \\
               SciTail &            TE &    23,596 &     2,126 &     1,304 \\
          NCBI-disease &           NER &     5,432 &       923 &       942 \\
                 BC2GM &           NER &    12,632 &     2,531 &     5,065 \\
          CHEMDNER-BIO &           NER &    30,884 &    30,841 &    26,561 \\
                BC5CDR &           NER &     4,560 &     4,581 &     4,797 \\
              Linnaeus &           NER &    12,004 &     4,086 &     7,181 \\
            JNLPBA-DNA &           NER &     4,699 &       552 &       622 \\
            JNLPBA-RNA &           NER &       721 &        89 &       102 \\
             JNLPBA-CT &           NER &     4,792 &       420 &     1,422 \\
             JNLPBA-CL &           NER &     2,596 &       284 &       377 \\
                AnatEM &           NER &     5,861 &     2,118 &     3,830 \\
                  AnEM &           NER &       164 &       137 &        30 \\
              BioInfer &           NER &       894 &         0 &       206 \\
           BioNLP-2009 &           NER &       756 &       260 &       150 \\
       BioNLP-2011-EPI &           NER &       600 &       200 &         0 \\
        BioNLP-2011-GE &           NER &       856 &         0 &       338 \\
        BioNLP-2011-ID &           NER &       151 &        46 &       117 \\
       BioNLP-2011-REL &           NER &       756 &       150 &       260 \\
        BioNLP-2013-CG &           NER &       300 &       100 &       200 \\
        BioNLP-2013-GE &           NER &       194 &       212 &       256 \\
       BioNLP-2013-GRO &           NER &       150 &        50 &       100 \\
        BioNLP-2013-PC &           NER &       260 &        90 &       175 \\
        BioNLP-2019-BB &           NER &       132 &        66 &         0 \\
                BioRED &           NER &       400 &       100 &       100 \\
              BioRelEx &           NER &     1,402 &       201 &         0 \\
            CellFinder &           NER &         5 &         0 &         5 \\
                 CHEBI &           NER &       476 &         0 &         0 \\
              CHEMDNER &           NER &     2,915 &     2,906 &     2,477 \\
              ChemProt &           NER &     1,020 &       612 &       800 \\
                  CHIA &           NER &     1,932 &         0 &         0 \\
                   CPI &           NER &     1,808 &         0 &         0 \\
                   DDI &           NER &       673 &         0 &       279 \\
              DrugProt &           NER &     3,500 &       750 &         0 \\
               EBM-NLP &           NER &     4,735 &         0 &       187 \\
                EU-ADR &           NER &       299 &         0 &         0 \\
               GENETAG &           NER &     3,875 &     1,311 &     2,567 \\
            PTM-Events &           NER &       112 &         0 &         0 \\
            GENIA-Term &           NER &     2,000 &         0 &         0 \\
             GNormPlus &           NER &       418 &         0 &       261 \\
                HPRD50 &           NER &        34 &         0 &         9 \\
           MedMentions &           NER &     2,635 &       878 &       879 \\
                 miRNA &           NER &       201 &         0 &       100 \\
                  MLEE &           NER &       130 &        44 &        87 \\
              NLM-Gene &           NER &       450 &         0 &       100 \\
              NLM-Chem &           NER &        80 &        20 &        50 \\
                OSIRIS &           NER &       105 &         0 &         0 \\
                   PDR &           NER &       179 &         0 &         0 \\
       PICO-Annotation &           NER &       361 &         0 &         0 \\
               ProGene &           NER &    20,055 &     1,109 &     2,414 \\
         SCAI-Chemical &           NER &        67 &         0 &         0 \\
          SCAI-Disease &           NER &       330 &         0 &         0 \\
                  SETH &           NER &       433 &         0 &         0 \\
               SPL-ADR &           NER &       101 &         0 &         0 \\
              tmVar-v1 &           NER &       213 &         0 &       101 \\
              tmVar-v2 &           NER &       158 &         0 &         0 \\
              tmVar-v3 &           NER &         0 &         0 &       493 \\
         Verspoor-2013 &           NER &       117 &         0 &         0 \\
             MedDialog &      TXTCLASS &       981 &       126 &       122 \\
               SciCite &      TXTCLASS &     8,243 &       916 &     1,861 \\
   Hallmarks-of-Cancer &      TXTCLASS &    12,119 &     1,798 &     3,547 \\
            GEOKhoj-v1 &      TXTCLASS &    25,000 &         0 &     5,000 \\
          BC7-LitCovid &      TXTCLASS &    24,960 &     2,500 &     6,239 \\
       AskAPatient-NED &           NED &    15,612 &       845 &       867 \\
            BC5CDR-NED &           NED &       500 &       500 &       500 \\
                Bio-ID &           NED &    11,366 &         0 &         0 \\
    BioNLP-2019-BB-NED &           NED &       132 &        66 &         0 \\
            BioRED-NED &           NED &       400 &       100 &       100 \\
          BioRelEx-NED &           NED &     1,402 &       201 &         0 \\
               CPI-NED &           NED &     1,808 &         0 &         0 \\
         GNormPlus-NED &           NED &       418 &         0 &       261 \\
          Linnaeus-NED &           NED &        95 &         0 &         0 \\
                 MeDAL &           NED & 3,000,000 & 1,000,000 & 1,000,000 \\
       MedMentions-NED &           NED &     2,635 &       878 &       879 \\
             miRNA-NED &           NED &       201 &         0 &       100 \\
          MuchMore-NED &           NED &     7,820 &         0 &         0 \\
      NCBI-disease-NED &           NED &       592 &       100 &       100 \\
          NLM-Gene-NED &           NED &       450 &         0 &       100 \\
          NLM-Chem-NED &           NED &        80 &        20 &        50 \\
            OSIRIS-NED &           NED &       105 &         0 &         0 \\
           SPL-ADR-NED &           NED &       101 &         0 &         0 \\
          tmVar-v2-NED &           NED &       158 &         0 &         0 \\
          tmVar-v3-NED &           NED &         0 &         0 &       493 \\
           TwADR-L-NED &           NED &     4,816 &       115 &       143 \\
               AnEM-RE &            RE &        22 &         5 &        13 \\
             BC5CDR-RE &            RE &       500 &       500 &       500 \\
           BioInfer-RE &            RE &       642 &         0 &       142 \\
    BioNLP-2011-REL-RE &            RE &       378 &        92 &         0 \\
     BioNLP-2013-GE-RE &            RE &        40 &        41 &         0 \\
    BioNLP-2013-GRO-RE &            RE &       149 &        48 &         0 \\
     BioNLP-2019-BB-RE &            RE &       121 &        59 &         0 \\
             BioRED-RE &            RE &       395 &        97 &       100 \\
           BioRelEx-RE &            RE &     1,263 &       178 &         0 \\
              CHEBI-RE &            RE &       415 &         0 &         0 \\
           ChemProt-RE &            RE &       767 &       443 &       620 \\
               CHIA-RE &            RE &     1,876 &         0 &         0 \\
                CPI-RE &            RE &     1,246 &         0 &         0 \\
                DDI-RE &            RE &       510 &         0 &       191 \\
           DrugProt-RE &            RE &     2,433 &       542 &         0 \\
             EU-ADR-RE &            RE &       253 &         0 &         0 \\
             HPRD50-RE &            RE &        28 &         0 &         8 \\
                  IEPA &            RE &       114 &         0 &        26 \\
                 LLL05 &            RE &        77 &         0 &         0 \\
               MLEE-RE &            RE &        32 &        11 &        16 \\
           MuchMore-RE &            RE &     7,734 &         0 &         0 \\
               SETH-RE &            RE &       212 &         0 &         0 \\
            SPL-ADR-RE &            RE &        96 &         0 &         0 \\
      Verspoor-2013-RE &            RE &       114 &         0 &         0 \\
            AnEM-COREF &         COREF &        10 &         2 &        14 \\
     BioNLP-2009-COREF &         COREF &       536 &       110 &         0 \\
 BioNLP-2011-EPI-COREF &         COREF &       440 &       168 &         0 \\
  BioNLP-2011-GE-COREF &         COREF &       571 &         0 &         0 \\
  BioNLP-2011-ID-COREF &         COREF &       170 &        31 &         0 \\
 BioNLP-2011-REL-COREF &         COREF &       535 &       110 &         0 \\
  BioNLP-2013-CG-COREF &         COREF &       466 &       176 &         0 \\
  BioNLP-2013-GE-COREF &         COREF &        53 &        41 &         0 \\
  BioNLP-2013-PC-COREF &         COREF &       455 &       128 &         0 \\
        BioRelEx-COREF &         COREF &     1,143 &       167 &         0 \\
      PTM-Events-COREF &         COREF &        25 &         0 &         0 \\
            MLEE-COREF &         COREF &       198 &        57 &       113 \\
             PDR-COREF &         COREF &        19 &         0 &         0 \\
           Bio-SimVerb &           STS &     1,000 &         0 &         0 \\
            Bio-SimLex &           STS &       988 &         0 &         0 \\
               BIOSSES &           STS &        64 &        16 &        20 \\
               EHR-Rel &           STS &     3,741 &         0 &         0 \\
               MayoSRS &           STS &       101 &         0 &         0 \\
                   MQP &           STS &     3,048 &         0 &         0 \\
                UMNSRS &           STS &     1,153 &         0 &         0 \\
        BioNLP-2009-EE &            EE &       695 &       150 &         0 \\
    BioNLP-2011-EPI-EE &            EE &       383 &       121 &         0 \\
     BioNLP-2011-GE-EE &            EE &       765 &         0 &         0 \\
     BioNLP-2011-ID-EE &            EE &       110 &        30 &         0 \\
     BioNLP-2013-CG-EE &            EE &       299 &       100 &         0 \\
     BioNLP-2013-GE-EE &            EE &       149 &       157 &         0 \\
     BioNLP-2013-PC-EE &            EE &       257 &        90 &         0 \\
         PTM-Events-EE &            EE &       111 &         0 &         0 \\
               MLEE-EE &            EE &       127 &        44 &        87 \\
                PDR-EE &            EE &       167 &         0 &         0 \\
       MuchMore-TRANSL &        TRANSL &     6,374 &         0 &         0 \\
               ParaMed &        TRANSL &    62,127 &     2,036 &     2,102 \\
                SciELO &        TRANSL & 3,006,699 &         0 &         0 \\
          Medical-Data & TEXTPAIRCLASS &     5,279 &         0 &         0 \\
                MeQSum &           SUM &     1,000 &         0 &         0 \\
        Multi-XScience &           SUM &    30,369 &     5,066 &     5,093 \\
\label{tab:dataset_collection}\end{longtable}

\end{document}